\documentclass{article} % For LaTeX2e
\usepackage{iclr2026_conference,times}
\usepackage[ruled,vlined,linesnumbered]{algorithm2e}
\usepackage{amsmath,amsfonts,bm}

\def\eqref#1{equation~\ref{#1}}
\def\1{\bm{1}}

\DeclareMathAlphabet{\mathsfit}{\encodingdefault}{\sfdefault}{m}{sl}
\SetMathAlphabet{\mathsfit}{bold}{\encodingdefault}{\sfdefault}{bx}{n}

\usepackage{tikz}
\usetikzlibrary{shapes.geometric, arrows.meta, positioning, fit, backgrounds}
\usepackage{hyperref}
\usepackage{url}
\usepackage{graphicx} % For images
\usepackage{amsmath}  % For math formatting
\usepackage{amssymb}  % For symbols like arrows
\usepackage[inkscapelatex=false]{svg}

\definecolor{myMagenta}{RGB}{200, 0, 200}
\definecolor{myOrange}{RGB}{200, 100, 0}
\definecolor{myCyan}{RGB}{0, 180, 200}
\definecolor{myViolet}{RGB}{100, 0, 200}
\usepackage{pgfplots}
\usepackage{subcaption} 
\pgfplotsset{compat=1.18}
\usepackage{booktabs}   % For professional looking tables
\usepackage{tabularx}   % For auto-adjusting column widths
\usepackage{xcolor}     % For text coloring
\usepackage{amsmath}    % For math symbols
\usepackage{geometry}   % To ensure margins allow for the wide table

\title{Continual learning and refinement of causal models through dynamic predicate invention}

\author{Enrique Crespo-Fernández, Oliver Ray, Telmo de Menezes e Silva Filho \& Peter Flach\\
University of Bristol\\
Bristol, UK \\
}
\iclrfinalcopy % Uncomment for camera-ready version, but NOT for submission.
\begin{document}

\maketitle
\vspace{-0.5cm}
\begin{abstract} 
Efficiently navigating complex environments requires agents to internalize the underlying logic of their world, yet standard world modelling methods often struggle with sample inefficiency, lack of transparency, and poor scalability. We propose a framework for constructing symbolic causal world models entirely online by integrating continuous model learning and repair into the agent's decision loop, by leveraging the power of Meta-Interpretive Learning and predicate invention to find semantically meaningful and reusable abstractions, allowing an agent to construct a hierarchy of disentangled, high-quality concepts from its observations. We demonstrate that our lifted inference approach scales to domains with complex relational dynamics, where propositional methods suffer from combinatorial explosion, while achieving sample-efficiency orders of magnitude higher than the established PPO neural-network-based baseline.
\end{abstract}
\begin{figure}[!b]
    \centering
    % --- COLUMN 1: IMAGE ---
    \begin{minipage}[t]{0.30\textwidth}
        \vspace{0pt}
        \includegraphics[width=\linewidth]{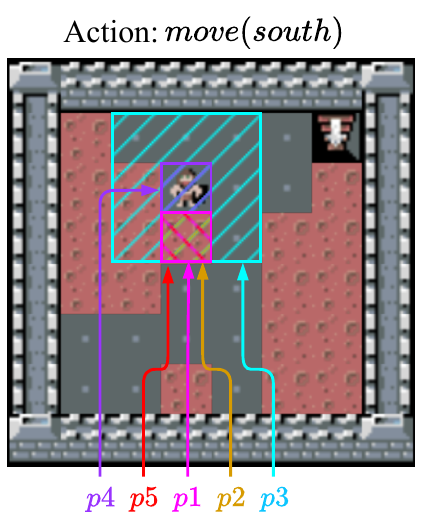}
    \end{minipage}\hfill
    %
    % --- COLUMN 2: DYNAMICS & CONSTRAINTS ---
    \begin{minipage}[t]{0.32\textwidth}
        \vspace{0pt}
        \scriptsize
        \textbf{Learnt Dynamics}
        \begin{align*}
            & \text{\itshape\color{gray} \% Rule: movement} \\
            & \mathrm{at}_{t+1}(A,B) \Leftarrow \textcolor{magenta}{p1}(A,B) \\[0.5em]
            & \text{\itshape\color{gray} \% Rule: dying} \\
            & \mathrm{dead}_{t+1}(A) \Leftarrow \textcolor{red}{p5}(A)
        \end{align*}
        
        \vspace{0.2em} 
        
        \textbf{Learnt Constraints}
        \begin{align*}
            & \text{\itshape\color{gray} \% Agent cannot occupy two locations} \\
            & \mathrm{at}_t(A,B) \otimes \mathrm{at}_t(A,D) \\[0.5em]
            & \text{\itshape\color{gray} \% A cannot be dead and alive} \\
            & \mathrm{alive}_t(A) \otimes \mathrm{dead}_t(A)
        \end{align*}
    \end{minipage}\hfill
    %
    % --- COLUMN 3: ABSTRACTIONS ---
    \begin{minipage}[t]{0.34\textwidth}
        \vspace{0pt}
        \scriptsize 
        \textbf{Learnt Abstractions}
        \begin{align*}
            & \text{\itshape\color{gray} \% Agent A alive at loc B} \\
            & \textcolor{violet}{p4}(A,B) \leftarrow \mathrm{at}(A,B) \wedge \mathrm{alive}(A) \\[0.5em]
            & \text{\itshape\color{gray} \% Surroundings of the agent A} \\
            & \textcolor{cyan}{p3}(A,C,D) \leftarrow \textcolor{violet}{p4}(A,B) \wedge \mathrm{adjacent}(B,C,D) \\[0.5em]
            & \text{\itshape\color{gray} \% Agent A moving towards cell B} \\
            & \textcolor{brown}{p2}(A,B) \leftarrow move(C) \wedge \textcolor{cyan}{p3}(A,C,B) \\[0.5em]
            & \text{\itshape\color{gray} \% Valid move} \\
            & \textcolor{magenta}{p1}(A,B) \leftarrow \textcolor{brown}{p2}(A,B) \wedge \mathrm{not\_wall}(B) \\[0.5em]
            & \text{\itshape\color{gray} \% Moving into lava} \\
            & \textcolor{red}{p5}(A) \leftarrow \textcolor{brown}{p2}(A,B) \wedge \mathrm{is\_lava}(B)
        \end{align*}
    \end{minipage}
    \caption{Visualization of a selected set of rules from the learnt symbolic causal model on the MiniHack `Lava Crossing' task. The environment is modelled as a hierarchy of interpretable concepts, transition rules, and physical constraints.
    (Left) State Interpretation: Colored overlays illustrate how the Learnt Abstractions (Right) ground to specific regions of the state space. The agent ($p4$, purple) senses its neighborhood ($p3$, cyan). The concept of "moving" ($p2$, orange) is reused in two rules: the one modelling movement ($p1$) and the one modelling death ($p5$).
    (Center) Dynamics \& Constraints: The Learnt Dynamics use these high-level abstractions to predict state evolution. For example, the dying rule is triggered only when the abstract condition $p5$ (moving into lava) is met. Learnt Constraints enforce physical consistency, such as mutual exclusion ($\otimes$), ensuring the agent cannot be simultaneously alive and dead or occupy multiple coordinates.}
    \label{fig:three_column_logic}
\end{figure}

\section{Introduction}
Intelligent agents deployed in dynamic environments must learn compact world models from scarce experience, reason with them, and adapt when predictions fail. Modern deep model-based reinforcement learning offers powerful function approximation but remains data-hungry and opaque \citep{suttonReinforcementLearningIntroduction2018,moerlandModelbasedReinforcementLearning2022, zhangAlphaZero2020}. Conversely, traditional symbolic systems possess structural properties—specifically, inherent compositionality, the capacity for arbitrary-depth computation, and interpretability—that make them ideal candidates for the backbone of a causal agent. However, classical automated planning typically presupposes a hand-engineered symbolic world model \citep{ghallabAutomatedPlanningTheory2004,fikesStripsNewApproach1971, yuExplainableReinforcementLearning2023, smetRelationalNeurosymbolicMarkov2025}. 

Bridging these paradigms requires a framework that acquires a symbolic dynamics model from experience while retaining the deliberative strengths of classical planning. The acquisition of such models requires an agent to formulate hypotheses that rationalize environmental changes as they occur, to test their consistency over time, and to update beliefs when they are falsified \citep{suttonReinforcementLearningIntroduction2018, evansMakingSenseSensory2021}. Crucially, if the aim is general-purpose agency, the learner must be able to formulate and refine these hypotheses during interaction without reliance on pre-collected trajectories, batched histories, or external oracles, as is common in many existing Inductive Logic Programming (ILP) approaches \citep{evansMakingSenseSensory2021,cropperInductiveGeneralGame2020,cropperLearningProgramsLearning2021}. To this end, we present a self-supervised, online framework for learning symbolic world models. Our agent operates in a continuous loop: it (i) incrementally \textbf{induces} a symbolic transition theory from streams of experience, (ii) \textbf{predicts} state evolution to plan actions, and (iii) \textbf{reacts} to prediction–observation mismatches by revising its model in real-time.

Building on recent developments in Meta-Interpretive Learning (MIL) \citep{muggletonMetainterpretiveLearningHigherorder2015,cropperLogicalReductionMetarules2020, patsantzisTopProgramConstruction2021}, our system generates logical explanations for unpredicted transitions in the language of First-Order Logic. We employ a template-based approach to manage the combinatorial complexity of the search space. Furthermore, we introduce a dynamic predicate invention that composes short, general abstract rules to define complex relationships. This hierarchical structure ensures that the learned dynamic rules remain concise, general, and interpretable while remaining sufficiently robust to capture deep causal dependencies.

The main contributions of this work can be summarized as follows. We introduce a \textbf{Continuous Model Repair} framework that utilizes prediction error to refine symbolic theories incrementally, eliminating the need for retraining. By using lifted inference and predicate invention, the complexity of our \textbf{Scale-Invariant Learning} mechanism depends only on the logic depth and vocabulary size and not on the grounding size, a standard limitation of propositional methods (ASP- and SAT-based). We empirically validate the \textbf{Sample Efficiency} of our system on grid-world environments, where it achieves superior sample efficiency compared to standard PPO \cite{schulmanProximalPolicyOptimization2017} baselines. 

By integrating symbolic learning and model repair into a single loop, we enable agents that are not only effective but also transparent and data-efficient. The remainder of this paper is structured as follows: Section 2 formalizes the MIL framework and the self-supervised Predict-Verify-Refine cycle. Section 3 presents an empirical evaluation of the system's sample efficiency and model repair capabilities in the MiniHack environment. Section 4 situates our work within the broader context of symbolic and neurosymbolic world models, and Section 5 concludes with a discussion of future research directions.

\section{Method}
This section details the MIL framework for acquiring predictive world models. We employ a self-supervised continuous learning paradigm in which the agent induces a logic program $H$ to predict state transitions and iteratively refines it based on prediction failures.

\subsection{Problem Definition}
The learning task is to induce a hypothesis $H$ that rationalizes transitions in a sequence $S = \langle S_0, S_1, \dots, S_n \rangle$ within a deterministic, fully observable environment. A state $S_t$ is a set of ground atoms. The goal is to learn a transition function $T: S \times A \rightarrow S'$ as a logic program $H = \langle Abs, Dyn, Con \rangle$. Here, $Abs$ is a set of definite clauses defining invented predicates, enabling compositional representation of spatial or semantic relations of the form $p_i \leftarrow Body$. $Dyn$ is a set of transition rules governing the addition of atoms, taking the form $add(P) \leftarrow Body$. $Con$ is a set of rules governing the removal of atoms, taking the form $del(P) \leftarrow Body$. $Body$ is a conjunction of literals drawn from the background knowledge $\mathcal{B}$, the current state $S_t$, and the set of invented predicates defined in $Abs$.

The transition logic is defined by:
$$
S_{t+1} = (S_t \setminus \mathcal{D}) \cup \mathcal{A}
$$
where $\mathcal{A} = \{h | H_{Dyn} \cup S_t \cup \mathcal{B} \models h\}$ and $\mathcal{D} = \{h | H_{Con} \cup S_t \cup \mathcal{B} \models h\}$, with $\mathcal{B}$ representing background knowledge (static predicates). We assume \textbf{lifted dynamics}, meaning rules depend on object relations rather than identities. The search space is the power set of the logic language $\mathcal{L}$. To manage the super-exponential complexity online, we use a \textbf{Predict-Verify-Refine} cycle rather than batch learning. We maintain a current hypothesis $H_t$; when $H_t$ fails to predict $S_{t+1}$, we trigger localized MIL search to repair the theory via generalization (metarule-guided construction) or specialization (clause pruning), see Algorithm \ref{alg:online_mil} in Appendix B.

\begin{algorithm}
\caption{Metarule-Guided Abduction with Predicate Reuse}
\label{alg:Hypothesis_construction}
\SetKwInOut{Input}{Input}
\SetKwInOut{Output}{Output}

% Use standard Function formatting to avoid 'Undefined control sequence'
\textbf{Function} \textit{MetaruleInduction}($Literal\ L, Depth\ k$): \\
\If{$k > D_{max}$}{
    \textbf{return} failure \tcp*{Terminate at max abstraction depth}
}
$\mathcal{H} \leftarrow \emptyset, \mathcal{T} \leftarrow \emptyset$\;

\ForEach{metarule $m \in \mathcal{M}$ ($H \leftarrow B_1, \dots, B_n$)}{
    Apply substitution $\theta$ such that $H\theta = L$ \;
    $Body \leftarrow [\ ]$\;
    
    \ForEach{literal $lit \in [B_1\theta, \dots, B_n\theta]$}{
        \eIf{$lit$ is primitive OR entails $\mathcal{B}$ }{
            $Body$.append($lit$)\;
        }{
            \tcp{Recursive abduction for predicate invention}
            $(P, Rule_p, Type_{p}) \leftarrow \textit{MetaruleInduction}(lit, k+1)$\;
            $Body$.append($P$)\ \tcp*{Invented predicate P added to the body}
            $\mathcal{H} \leftarrow \mathcal{H} \cup Rule_{p}$\ \tcp*{Rule that explains P is stored}
            $\mathcal{T} \leftarrow \mathcal{T} \cup Type_{p}$ \tcp*{Type signature of P is stored}
        }
    }
    \uIf{$k == 0$}{
        $\mathcal{H} \leftarrow \mathcal{H} \cup \{H \leftarrow Body\}$\;
        $P_{ret} \leftarrow L$\;
    }
    \Else{
        $P_{ret} \leftarrow \textit{ReuseOrRegister}(Body, \mathcal{H})$\;
    }
}
\textbf{return} $(P_{ret}, \mathcal{H}, \mathcal{T})$\;
\end{algorithm}

\subsection{Metarule-Guided Hypothesis construction and refinement}

To overcome the combinatorial explosion typical of inductive synthesis, we restrict the search space using
rule templates \citep{muggletonMetainterpretiveLearningHigherorder2015}. Rather than searching the space of all Horn clauses, we search the space of
proofs generated by a set of second-order templates called \textbf{metarules} $\mathcal{M}$. 

A metarule $M \in \mathcal{M}$ is a second-order clause defining a valid syntactic structure for a rule. For example, the Chain metarule allows the agent to discover transitive relationships, and the absorption metarule to restrict a relation with a property:
\begin{align*}
M_{chain}: P(X,Y) \leftarrow Q(X,Z), R(Z,Y) \\
M_{absorption}: P(X,Y) \leftarrow Q(X,Y),R(Y)
\end{align*}

While users select metarules, these typically have a general structure and can be reused across domains. Additionally, the system can compensate for a missing metarule by composing the available metarules through predicate invention.

\textbf{Top Program Construction.}
Building on the Louise system \citep{patsantzisTopProgramConstruction2021}, we avoid searching for a single hypothesis directly. Instead, we construct the \textbf{Top Program} $\top$, the most general logic program entailing observed positive examples ($E^+$) within constraints:
\begin{equation}
\top = \{ M\theta \mid M \in \mathcal{M}, \exists (E^+_i \in E^+) : \text{entails}(M\theta, E^+_i) \}
\end{equation}
In our online setting, $\top$ represents the set of all plausible explanations. Learning reduces to generalizing $\top$ to explain new changes and specializing $\top$ by pruning falsified clauses (see Figure \ref{fig:lattice} in Appendix A).

\textbf{Recursive Abduction and Predicate Invention.}
Hypotheses are constructed via abductive instantiation. Given an unexplained observation $O$, the system identifies a metarule $M$ and a substitution $\theta$ such that $head(M)\theta = O$. Body literals are resolved against $\mathcal{B} \cup S_t$ or further abduced. These unexplained literals become targets for further abduction using abstraction metarules $\mathcal{M}_{abs}$ (see Algorithm \ref{alg:Hypothesis_construction}. This process builds a hierarchical derivation chain in which higher-level predicates are defined in terms of lower-level ones, continuing recursively until the chain is grounded in $\mathcal{B}$ or the maximum depth $D_{max}$ is reached. To bound complexity, we impose a type system $\mathcal{T}$, considering only instantiations that respect predicate typing in $\mathcal{B}$. The user manually assigns a type to primary predicates; invented predicates inherit the type from the predicates in their body.

\textbf{Error Signals and Refinement.}
For any state transition $(S_t \rightarrow S_{t+1})$, the learner derives error signals via set difference operations on the observed states. We distinguish between two types of prediction errors that drive the lattice search, see Figure \ref{fig:lattice} in Appendix A:
\begin{itemize}
    \item \textbf{False Negatives ($FN$):} These represent observed changes in the environment that the current hypothesis $H$ failed to predict (i.e., $S_{t+1} \setminus S_{predicted}$). These examples trigger the \textbf{generalization} of $\top$ by invoking the abductive engine to generate new metarule instantiations that entail the missing atoms, adding them to $\top$ and $H$.
    \item \textbf{False Positives ($FP$):} These represent changes predicted by $H$ that did not occur in reality (i.e., $S_{predicted} \setminus S_{t+1}$). When the model hallucinates an effect, the system identifies the specific clauses in $H$ responsible for the hallucination and \textbf{specialises} $\top$ by pruning them. 
\end{itemize}
Explicitly constructing the set of all non-changes in a large environment is intractable. To mitigate it without intractable global checks, we employ an Inertia Assumption \citep{fikesStripsNewApproach1971}: the state is assumed invariant unless a specific rule in $H$ predicts a change. This reduces the verification step to checking only predicted effects ($S_{predicted} \setminus S_{t+1}$).

\subsection{Complexity Analysis}
\label{sec:complexity}

A defining characteristic of our framework is its scale invariance: the computational cost of hypothesis generation depends on the complexity of the underlying logic $\mathcal{D}_{max}$ rather than on the size of the state space. By operating on lifted predicates with intensional background knowledge, our system evaluates rules such as $adjacent/3$ via procedural attachment rather than by scanning a state matrix that grows exponentially with grid size.

While the search space for logic programs is theoretically infinite, our approach bounds the search through three structural constraints: Fixed Metarules, a Strong Type System, and Canonicalization. The worst-case complexity of expanding the hypothesis space for a single generalization step can be stated as:
\begin{equation}
    O(|\mathcal{M}| \cdot |\mathcal{P}_{typed}|^k \cdot D_{max})
\end{equation}
Here, $|\mathcal{M}|$ is the finite set of second-order metarules. $|\mathcal{P}_{typed}|$ is the effective branching factor, constrained by the type system. $k$ is the arity of the metarules (typically $k=2$). 

Our algorithm employs Top Program Construction with memorization, which effectively converts the search from a Tree Traversal into a \textbf{Directed Acyclic Graph Construction}. We maintain a \textbf{Global Predicate Registry} ($\Omega$) of all discovered predicate signatures. When the abductive engine generates a new rule body (e.g., $P \leftarrow Q, R$), it computes a canonical hash of the body literals. If a semantically equivalent predicate already exists in $\Omega$, the system reuses it rather than inventing a duplicate. We cache the results of abductive queries. If the system encounters a subgoal that has already been solved within the current "budget" of depth, it retrieves the solution in $O(1)$ time. By collapsing redundant branches and preventing the re-derivation of known concepts, the total work becomes the \textit{sum} of the work required to construct each layer of the hierarchy, rather than the product. 

\section{Experiments}
We evaluate our framework against three core desiderata for concept learning: the ability to perform continuous model repair; sample efficiency relative to neural baselines; and the robustness of the learned abstractions across varying domain sizes and semantic alignment.

% LaTeX code for MIL vs PPO comparison figure
% Requires: \usepackage{pgfplots}
% Add to preamble: \pgfplotsset{compat=1.18}

\begin{figure}[htbp]
\centering
\begin{minipage}[b]{0.48\textwidth}
    \centering
    \begin{tikzpicture}
    \begin{axis}[
        title={Model Size Evolution},
        title style={font=\small},
        width=\textwidth,
        height=0.75\textwidth,
        xlabel={Timestep},
        ylabel={Number of Clauses},
        xmin=0, xmax=100,
        ymin=0, ymax=150,
        legend pos=north east,
        legend style={font=\scriptsize},
        grid=major,
        grid style={gray!30},
        stack plots=y,
        area style,
    ]

    % Constraints (bottom layer)
    \addplot[fill=blue!30, draw=blue!50] coordinates {
        (1,0) (2,0) (3,0) (4,0) (5,0) (6,0) (7,1) (8,1) (9,1) (10,1)
        (11,2) (12,2) (13,2) (14,2) (15,2) (16,2) (17,2) (18,2) (19,2) (20,2)
        (21,2) (22,2) (23,2) (24,2) (25,2) (26,2) (27,2) (28,2) (29,2) (30,2)
        (31,2) (32,2) (33,2) (34,2) (35,2) (36,2) (37,2) (38,2) (39,2) (40,2)
        (41,2) (42,2) (43,2) (44,2) (45,2) (46,2) (47,2) (48,2) (49,2) (50,2)
        (51,2) (52,2) (53,2) (54,2) (55,2) (56,2) (57,2) (58,2) (59,2) (60,2)
        (61,2) (62,2) (63,2) (64,2) (65,2) (66,2) (67,2) (68,2) (69,2) (70,2)
        (71,2) (72,2) (73,2) (74,2) (75,2) (76,2) (77,2) (78,2) (79,2) (80,2)
        (81,2) (82,2) (83,2) (84,2) (85,2) (86,2) (87,2) (88,2) (89,2) (90,2)
        (91,2) (92,2) (93,2) (94,2) (95,2) (96,2) (97,2) (98,2) (99,2) (100,2)
    } \closedcycle;
    \addlegendentry{Constraints}

    % Dynamics (middle layer)
    \addplot[fill=green!40, draw=green!60] coordinates {
        (1,0) (2,0) (3,0) (4,0) (5,0) (6,0) (7,55) (8,14) (9,14) (10,14)
        (11,55) (12,35) (13,35) (14,35) (15,35) (16,35) (17,35) (18,35) (19,35) (20,35)
        (21,35) (22,13) (23,13) (24,13) (25,13) (26,13) (27,13) (28,13) (29,13) (30,13)
        (31,13) (32,13) (33,13) (34,13) (35,13) (36,13) (37,13) (38,13) (39,13) (40,13)
        (41,13) (42,13) (43,13) (44,13) (45,13) (46,13) (47,13) (48,13) (49,13) (50,13)
        (51,13) (52,13) (53,13) (54,13) (55,13) (56,13) (57,13) (58,13) (59,13) (60,13)
        (61,13) (62,13) (63,13) (64,13) (65,13) (66,13) (67,13) (68,13) (69,13) (70,13)
        (71,13) (72,13) (73,13) (74,13) (75,13) (76,13) (77,13) (78,13) (79,13) (80,13)
        (81,13) (82,13) (83,13) (84,13) (85,13) (86,13) (87,13) (88,13) (89,13) (90,13)
        (91,13) (92,13) (93,13) (94,13) (95,13) (96,13) (97,13) (98,13) (99,13) (100,13)
    } \closedcycle;
    \addlegendentry{Dynamics}

    % Abstractions (top layer)
    \addplot[fill=red!40, draw=red!60] coordinates {
        (1,0) (2,0) (3,0) (4,0) (5,0) (6,0) (7,76) (8,28) (9,28) (10,28)
        (11,83) (12,58) (13,58) (14,58) (15,58) (16,58) (17,58) (18,58) (19,58) (20,58)
        (21,58) (22,28) (23,28) (24,28) (25,28) (26,28) (27,28) (28,28) (29,28) (30,28)
        (31,28) (32,28) (33,28) (34,28) (35,28) (36,28) (37,28) (38,28) (39,28) (40,28)
        (41,28) (42,28) (43,28) (44,28) (45,28) (46,28) (47,28) (48,28) (49,28) (50,28)
        (51,28) (52,28) (53,28) (54,28) (55,28) (56,28) (57,28) (58,28) (59,28) (60,28)
        (61,28) (62,28) (63,28) (64,28) (65,28) (66,28) (67,28) (68,28) (69,28) (70,28)
        (71,28) (72,28) (73,28) (74,28) (75,28) (76,28) (77,28) (78,28) (79,28) (80,28)
        (81,28) (82,28) (83,28) (84,28) (85,28) (86,28) (87,28) (88,28) (89,28) (90,28)
        (91,28) (92,28) (93,28) (94,28) (95,28) (96,28) (97,28) (98,28) (99,28) (100,28)
    } \closedcycle;
    \addlegendentry{Abstractions}

    \end{axis}
    \end{tikzpicture}
\end{minipage}
\hfill
\begin{minipage}[b]{0.48\textwidth}
    \centering
    \begin{tikzpicture}
    \begin{axis}[
        title={Learning Curves},
        title style={font=\small},
        width=\textwidth,
        height=0.75\textwidth,
        xlabel={Episode},
        ylabel={Reward (20-ep moving avg)},
        xmin=0, xmax=500,
        ymin=-0.1, ymax=1.1,
        legend pos=south east,
        legend style={font=\scriptsize},
        grid=major,
        grid style={gray!30},
    ]

    % PPO moving average (computed from actual data)
    \addplot[blue, thick] coordinates {
        (20,0.0) (40,0.0) (60,0.0) (80,0.0) (100,0.0) (120,0.0)
        (140,0.025) (160,0.028) (180,0.0) (200,0.0) (220,0.0) (240,0.0)
        (260,0.036) (280,0.0) (300,0.037) (320,0.0) (340,0.070)
        (360,0.129) (380,0.227) (400,0.299) (420,0.352) (440,0.386)
        (460,0.470) (480,0.585) (500,0.640)
    };
    \addlegendentry{PPO}

    % MIL symbolic (100% success after episode 2)
    \addplot[red, thick, dashed] coordinates {
       (1,0.0) (2,0.63) (10,0.82) (20,0.82) (30,0.82) (40,0.82) (50,0.82)
        (60,0.82) (70,0.82) (80,0.82) (90,0.82) (500,0.82)
    };
    \addlegendentry{Ours}

    % First success markers
    % MIL first success: episode 2
    \addplot[only marks, mark=star, mark size=5pt, red, mark options={fill=red}]
        coordinates {(2, 0.63)};
    \node[anchor=south west, font=\small, red] at (axis cs:8,0.63) {First success ep. 2};

    % PPO first success: episode 129
    \addplot[only marks, mark=star, mark size=5pt, blue, mark options={fill=blue}]
        coordinates {(129, 0.025)};
    \node[anchor=north west, font=\small, blue] at (axis cs:135,0.2) {ep. 128};

    \end{axis}
    \end{tikzpicture}
\end{minipage}

\caption{Our system vs PPO on the 10$\times$10 grid version of the MiniHack `Lava Crossing' task. (a) Our system converges to 43 clauses (28 abstractions, 13 dynamics, 2 constraints) by step 23. (b) Our system reaches the goal at episode 2 since then it is able to consistently navigate the environment to it relying on it learn model; PPO requires 129 episodes for first success and it is not until episode 300 that it starts to converge.}
\label{fig:mil_ppo_comparison}
\end{figure}
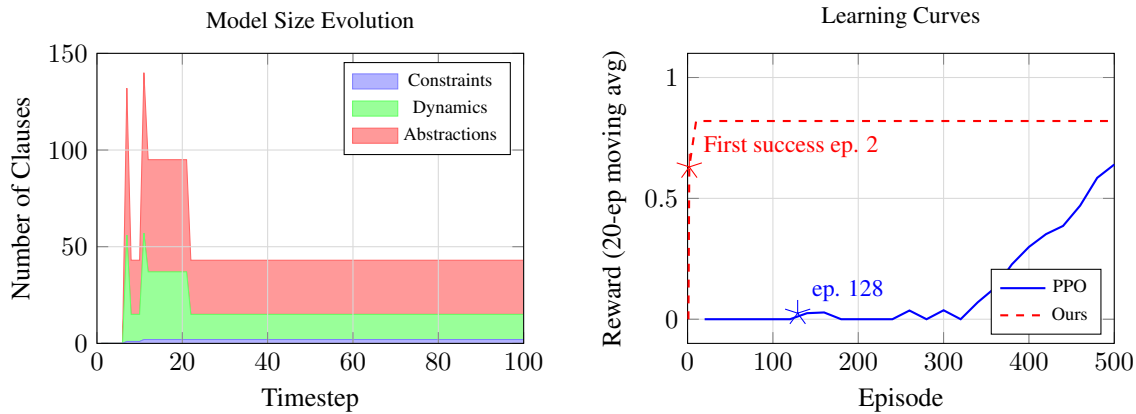

\subsection{Online Model Repair and Convergence}
To validate the Predict-Verify-Refine cycle, we track the evolution of the symbolic hypothesis $H$ during a single run in the MiniHack environment. Figure \ref{fig:mil_ppo_comparison} visualizes the internal model evolution. We monitor Model Size $|Dyn \cup Cons \cup Abs|$. Spikes in model size correspond to generalization events, where unexpected observations $FN$ trigger the invention of new candidate rules. Subsequent sharp drops correspond to specialization, where the agent prunes rules that lead to $FP$. By time step 20, the model stabilizes into a set of plausible rules that have not been disproven. Prediction error drops to zero, demonstrating that the system successfully disentangles valid causal mechanisms from transient noise. 

\subsection{Sample Efficiency and Scale Invariance}

We compare the sample efficiency of our Online MIL agent with that of a standard PPO baseline initialized with pre-trained CNN weights. As shown in Figure \ref{fig:mil_ppo_comparison}, the symbolic agent solves the environment in Episode 2 (one-shot learning after a single failure), whereas PPO requires approx. 128 episodes.

Although a logic-based learner is expected to outperform a gradient-based learner on grid worlds, this result highlights the representational efficiency of lifted concepts. Unlike neural weights, which approximate the state manifold, our agent immediately lifts observations into a relational space. Consequently, the "danger" concept learned on a $10 \times 10$ grid is structurally identical to one on a $100 \times 100$ grid. We confirmed that the model learned in the small environment generalizes zero-shot to the larger grid, demonstrating that the learned concepts are invariant to the state-space scale.

\subsection{Semantic alignment and Interpretability}
A key advantage of our approach over neural world models is the immediate interpretability of the latent space. The system invents predicates ($p_n$) to compress the state space. Figure \ref{fig:three_column_logic} illustrates the semantic alignment between the automatically invented predicates and human concepts for the 'Lava Crossing' task. The predicate $p_2$, for example, is reused to define the death and movement transition rules, demonstrating that the system creates hierarchical abstractions rather than flat correlations.

\section{Related Work}

\textbf{Symbolic and Neurosymbolic World Models.}
Learning symbolic dynamics is traditionally framed as batch constraint satisfaction. While the Apperception Engine \citep{evansMakingSenseSensory2021} offers robust unification, however it is framed as a constraint satisfaction problem, which makes it computationally prohibitive for real-time agents in big environments. Neurosymbolic alternatives \citep{athalyePixelsPredicatesLearning2025, silverLearningSymbolicOperators2021, silverLearningNeuroSymbolicSkills2023,piriyakulkijPoEWorldCompositionalWorld2025} improve speed but often rely on expensive pre-trained LLMs, which poses a challenge for systems that work offline or are limited on compute. Our framework bridges this gap by framing the problem as local, iterative refinement, updating beliefs only when immediate predictions are falsified. Our work system exclusively relies on pure logic programming rather than gradient-based search. This ensures the automated acquisition of the transition model results in a transparent, verifiable First-Order Logic theory at every step.

\textbf{Non-Monotonic Learning.}
Modern Inductive learners such as FastLAS \citep{lawFastLASScalableInductive2020} and Popper \citep{cropperLearningProgramsLearning2021, cropperInductiveGeneralGame2020} typically operate in a batch setting, requiring full traces or labelled examples. We extend the Lousie  \cite{patsantzisTopProgramConstruction2021} system by enabling incremental repair of transition rules during exploration via a self-supervision framework and by maintaining a persistent predicate registry, we enable the incremental repair of transition rules during exploration. Recently, PyGol \citep{vargheseOneShotLearningAutonomous2025} outperformed Deep RL in sample efficiency, taking a similar approach to ours. However, they focused on learning policies and are not able to perform predicate invention. 

\textbf{Abstraction and Program Synthesis.}
Systems like DreamCoder \citep{ellisDreamCoderBootstrappingInductive2021} compress solution spaces via offline ``wake-sleep'' cycles, creating a lag between observation and adaptation. In contrast, our system performs abstraction online. By inventing reusable predicates \textit{during} the decision loop, we merge the wake and sleep phases, making hierarchical concepts immediately available for planning.

\section{Conclusion}
This work introduces a novel online learning framework that leverages the representational power of invented predicates to generate compact, interpretable transition functions online. We show that this system is effective at learning models of classic RL environments. Additionally, we show that, when combined with an off-the-shelf planning algorithm, the agent efficiently solves the MiniHack lava environment, surpassing established RL benchmarks. These results suggest that this approach opens avenues for research into methods to address longstanding RL challenges in which agents must interact with diverse objects and reason about their behaviour. A promising research direction is to enforce active model refinement, in which the agent sets the rule body as a goal and tests whether the rule is valid. Other promising avenues of research include moving towards probabilistic logic and the use of neural predicates \citep{manhaeveNeuralProbabilisticLogic2021, evansMakingSenseRaw2021, smetRelationalNeurosymbolicMarkov2025}. This would allow a transition from hand-engineered domain encoding to a more autonomous learning framework in which primary predicates are also learnt. Combining this method with a neural policy could enable the agent to plan and react efficiently and robustly.

\bibliography{references}

@article{muggletonMetainterpretiveLearningHigherorder2015,
	title = {Meta-interpretive learning of higher-order dyadic datalog: predicate invention revisited},
	volume = {100},
	issn = {0885-6125, 1573-0565},
	shorttitle = {Meta-interpretive learning of higher-order dyadic datalog},
	url = {http://link.springer.com/10.1007/s10994-014-5471-y},
	doi = {10.1007/s10994-014-5471-y},
	language = {en},
	number = {1},
	urldate = {2026-01-30},
	journal = {Machine Learning},
	author = {Muggleton, Stephen H. and Lin, Dianhuan and Tamaddoni-Nezhad, Alireza},
	month = jul,
	year = {2015},
	pages = {49--73},
}

@article{smetRelationalNeurosymbolicMarkov2025,
	title = {Relational {Neurosymbolic} {Markov} {Models}},
	volume = {39},
	copyright = {Copyright (c) 2025 Association for the Advancement of Artificial Intelligence},
	issn = {2374-3468},
	url = {https://ojs.aaai.org/index.php/AAAI/article/view/33777},
	doi = {10.1609/aaai.v39i15.33777},
	language = {en},
	number = {15},
	urldate = {2026-01-30},
	journal = {Proceedings of the AAAI Conference on Artificial Intelligence},
	author = {Smet, Lennert De and Venturato, Gabriele and Raedt, Luc De and Marra, Giuseppe},
	month = apr,
	year = {2025},
	pages = {16181--16189},
}

@article{cropperLogicalReductionMetarules2020,
	title = {Logical reduction of metarules},
	volume = {109},
	issn = {0885-6125, 1573-0565},
	url = {http://link.springer.com/10.1007/s10994-019-05834-x},
	doi = {10.1007/s10994-019-05834-x},
	language = {en},
	number = {7},
	urldate = {2026-01-30},
	journal = {Machine Learning},
	author = {Cropper, Andrew and Tourret, Sophie},
	month = jul,
	year = {2020},
	pages = {1323--1369},
}

@article{manhaeveNeuralProbabilisticLogic2021,
	title = {Neural probabilistic logic programming in {DeepProbLog}},
	volume = {298},
	issn = {0004-3702},
	url = {https://www.sciencedirect.com/science/article/pii/S0004370221000552},
	doi = {10.1016/j.artint.2021.103504},
	urldate = {2026-01-30},
	journal = {Artificial Intelligence},
	author = {Manhaeve, Robin and Dumančić, Sebastijan and Kimmig, Angelika and Demeester, Thomas and De Raedt, Luc},
	month = sep,
	year = {2021},
	pages = {103504},
}

@misc{athalyePixelsPredicatesLearning2025,
	title = {From {Pixels} to {Predicates}: {Learning} {Symbolic} {World} {Models} via {Pretrained} {Vision}-{Language} {Models}},
	shorttitle = {From {Pixels} to {Predicates}},
	url = {http://arxiv.org/abs/2501.00296},
	doi = {10.48550/arXiv.2501.00296},
	urldate = {2026-01-30},
	publisher = {arXiv},
	author = {Athalye, Ashay and Kumar, Nishanth and Silver, Tom and Liang, Yichao and Wang, Jiuguang and Lozano-Pérez, Tomás and Kaelbling, Leslie Pack},
	month = jun,
	year = {2025},
	note = {arXiv:2501.00296 [cs]},
}

@article{lawFastLASScalableInductive2020,
	title = {{FastLAS}: {Scalable} {Inductive} {Logic} {Programming} {Incorporating} {Domain}-{Specific} {Optimisation} {Criteria}},
	volume = {34},
	copyright = {Copyright (c) 2020 Association for the Advancement of Artificial Intelligence},
	issn = {2374-3468},
	shorttitle = {{FastLAS}},
	url = {https://ojs.aaai.org/index.php/AAAI/article/view/5678},
	doi = {10.1609/aaai.v34i03.5678},
	language = {en},
	number = {03},
	urldate = {2026-01-30},
	journal = {Proceedings of the AAAI Conference on Artificial Intelligence},
	author = {Law, Mark and Russo, Alessandra and Bertino, Elisa and Broda, Krysia and Lobo, Jorge},
	month = apr,
	year = {2020},
	pages = {2877--2885},
}

@article{cropperInductiveGeneralGame2020,
	title = {Inductive general game playing},
	volume = {109},
	issn = {1573-0565},
	url = {https://doi.org/10.1007/s10994-019-05843-w},
	doi = {10.1007/s10994-019-05843-w},
	language = {en},
	number = {7},
	urldate = {2026-01-30},
	journal = {Machine Learning},
	author = {Cropper, Andrew and Evans, Richard and Law, Mark},
	month = jul,
	year = {2020},
	pages = {1393--1434},
}

@inproceedings{vargheseOneShotLearningAutonomous2025,
	address = {Berlin, Heidelberg},
	title = {One-{Shot} {Learning} of {Autonomous} {Behaviour}: {A} {Meta} {Inverse} {Entailment} {Approach}},
	isbn = {978-3-032-09086-7},
	shorttitle = {One-{Shot} {Learning} of {Autonomous} {Behaviour}},
	url = {https://doi.org/10.1007/978-3-032-09087-4_4},
	doi = {10.1007/978-3-032-09087-4_4},
	urldate = {2026-01-30},
	booktitle = {Learning and {Reasoning}: 4th {International} {Joint} {Conference} on {Learning} and {Reasoning}, {IJCLR} 2024, and 33rd {International} {Conference} on {Inductive} {Logic} {Programming}, {ILP} 2024, {Nanjing}, {China}, {September} 20–22, 2024, {Proceedings}},
	publisher = {Springer-Verlag},
	author = {Varghese, Dany and Cyrus, Daniel and Patsantzis, Stassa and Trewern, James and Treloar, Alfie Anthony and Hunter, Alan and Tamaddoni-Nezhad, Alireza},
	month = nov,
	year = {2025},
	pages = {48--65},
}

@article{patsantzisTopProgramConstruction2021,
	title = {Top program construction and reduction for polynomial time {Meta}-{Interpretive} learning},
	volume = {110},
	issn = {1573-0565},
	url = {https://doi.org/10.1007/s10994-020-05945-w},
	doi = {10.1007/s10994-020-05945-w},
	language = {en},
	number = {4},
	urldate = {2026-01-20},
	journal = {Machine Learning},
	author = {Patsantzis, S. and Muggleton, S. H.},
	month = apr,
	year = {2021},
	pages = {755--778},
}

@misc{moerlandModelbasedReinforcementLearning2022,
	title = {Model-based {Reinforcement} {Learning}: {A} {Survey}},
	shorttitle = {Model-based {Reinforcement} {Learning}},
	url = {http://arxiv.org/abs/2006.16712},
	doi = {10.48550/arXiv.2006.16712},
	urldate = {2025-08-29},
	publisher = {arXiv},
	author = {Moerland, Thomas M. and Broekens, Joost and Plaat, Aske and Jonker, Catholijn M.},
	month = mar,
	year = {2022},
	note = {arXiv:2006.16712 [cs]},
}

@book{ghallabAutomatedPlanningTheory2004,
	title = {Automated {Planning}: {Theory} and {Practice}},
	isbn = {978-0-08-049051-9},
	shorttitle = {Automated {Planning}},
	language = {en},
	publisher = {Elsevier},
	author = {Ghallab, Malik and Nau, Dana and Traverso, Paolo},
	month = may,
	year = {2004},
	note = {Google-Books-ID: uYnpze57MSgC},
}

@inproceedings{yuExplainableReinforcementLearning2023,
	address = {Macau, SAR China},
	title = {Explainable {Reinforcement} {Learning} via a {Causal} {World} {Model}},
	url = {https://www.ijcai.org/proceedings/2023/505},
	doi = {10.24963/ijcai.2023/505},
	language = {en},
	urldate = {2025-07-22},
	booktitle = {Proceedings of the {Thirty}-{Second} {International} {Joint} {Conference} on {Artificial} {Intelligence}},
	publisher = {International Joint Conferences on Artificial Intelligence Organization},
	author = {Yu, Zhongwei and Ruan, Jingqing and Xing, Dengpeng},
	month = aug,
	year = {2023},
	pages = {4540--4548},
}

@misc{piriyakulkijPoEWorldCompositionalWorld2025,
	title = {{PoE}-{World}: {Compositional} {World} {Modeling} with {Products} of {Programmatic} {Experts}},
	shorttitle = {{PoE}-{World}},
	url = {http://arxiv.org/abs/2505.10819},
	doi = {10.48550/arXiv.2505.10819},
	urldate = {2025-06-14},
	publisher = {arXiv},
	author = {Piriyakulkij, Wasu Top and Liang, Yichao and Tang, Hao and Weller, Adrian and Kryven, Marta and Ellis, Kevin},
	month = may,
	year = {2025},
	note = {arXiv:2505.10819 [cs]},
}

@article{fikesStripsNewApproach1971,
	title = {Strips: {A} new approach to the application of theorem proving to problem solving},
	volume = {2},
	issn = {0004-3702},
	shorttitle = {Strips},
	url = {https://www.sciencedirect.com/science/article/pii/0004370271900105},
	doi = {10.1016/0004-3702(71)90010-5},
	number = {3},
	urldate = {2025-04-04},
	journal = {Artificial Intelligence},
	author = {Fikes, Richard E. and Nilsson, Nils J.},
	month = dec,
	year = {1971},
	pages = {189--208},
}

@inproceedings{silverLearningNeuroSymbolicSkills2023,
	title = {Learning {Neuro}-{Symbolic} {Skills} for {Bilevel} {Planning}},
	issn = {2640-3498},
	url = {https://proceedings.mlr.press/v205/silver23a.html},
	language = {en},
	urldate = {2025-03-21},
	booktitle = {Proceedings of {The} 6th {Conference} on {Robot} {Learning}},
	publisher = {PMLR},
	author = {Silver, Tom and Athalye, Ashay and Tenenbaum, Joshua B. and Lozano-Pérez, Tomás and Kaelbling, Leslie Pack},
	month = mar,
	year = {2023},
	pages = {701--714},
}

@inproceedings{silverLearningSymbolicOperators2021,
	title = {Learning {Symbolic} {Operators} for {Task} and {Motion} {Planning}},
	issn = {2153-0866},
	url = {https://ieeexplore.ieee.org/document/9635941/?arnumber=9635941},
	doi = {10.1109/IROS51168.2021.9635941},
	urldate = {2025-03-21},
	booktitle = {2021 {IEEE}/{RSJ} {International} {Conference} on {Intelligent} {Robots} and {Systems} ({IROS})},
	author = {Silver, Tom and Chitnis, Rohan and Tenenbaum, Joshua and Kaelbling, Leslie Pack and Lozano-Pérez, Tomás},
	month = sep,
	year = {2021},
	pages = {3182--3189},
}

@article{evansMakingSenseRaw2021,
	title = {Making sense of raw input},
	volume = {299},
	issn = {0004-3702},
	url = {https://www.sciencedirect.com/science/article/pii/S0004370221000722},
	doi = {10.1016/j.artint.2021.103521},
	urldate = {2025-03-19},
	journal = {Artificial Intelligence},
	author = {Evans, Richard and Bošnjak, Matko and Buesing, Lars and Ellis, Kevin and Pfau, David and Kohli, Pushmeet and Sergot, Marek},
	month = oct,
	year = {2021},
	pages = {103521},
}

@book{suttonReinforcementLearningIntroduction2018,
	address = {Cambridge, MA, USA},
	title = {Reinforcement {Learning}: {An} {Introduction}},
	isbn = {978-0-262-03924-6},
	shorttitle = {Reinforcement {Learning}},
	publisher = {A Bradford Book},
	author = {Sutton, Richard S. and Barto, Andrew G.},
	month = oct,
	year = {2018},
}

@article{evansMakingSenseSensory2021,
	title = {Making sense of sensory input},
	volume = {293},
	issn = {0004-3702},
	url = {https://www.sciencedirect.com/science/article/pii/S0004370220301855},
	doi = {10.1016/j.artint.2020.103438},
	urldate = {2025-02-26},
	journal = {Artificial Intelligence},
	author = {Evans, Richard and Hernández-Orallo, José and Welbl, Johannes and Kohli, Pushmeet and Sergot, Marek},
	month = apr,
	year = {2021},
	pages = {103438},
}

@misc{schulmanProximalPolicyOptimization2017,
	title = {Proximal {Policy} {Optimization} {Algorithms}},
	url = {http://arxiv.org/abs/1707.06347},
	urldate = {2024-11-11},
	publisher = {arXiv},
	author = {Schulman, John and Wolski, Filip and Dhariwal, Prafulla and Radford, Alec and Klimov, Oleg},
	month = aug,
	year = {2017},
	note = {arXiv:1707.06347},
}

@incollection{zhangAlphaZero2020,
	address = {Singapore},
	title = {{AlphaZero}},
	isbn = {978-981-15-4095-0},
	url = {https://doi.org/10.1007/978-981-15-4095-0_15},
	doi = {10.1007/978-981-15-4095-0_15},
	language = {en},
	urldate = {2024-11-11},
	booktitle = {Deep {Reinforcement} {Learning}: {Fundamentals}, {Research} and {Applications}},
	publisher = {Springer},
	author = {Zhang, Hongming and Yu, Tianyang},
	editor = {Dong, Hao and Ding, Zihan and Zhang, Shanghang},
	year = {2020},
	pages = {391--415},
}

@article{cropperLearningProgramsLearning2021,
	title = {Learning programs by learning from failures},
	volume = {110},
	issn = {0885-6125, 1573-0565},
	url = {https://link.springer.com/10.1007/s10994-020-05934-z},
	doi = {10.1007/s10994-020-05934-z},
	language = {en},
	number = {4},
	urldate = {2024-10-02},
	journal = {Machine Learning},
	author = {Cropper, Andrew and Morel, Rolf},
	month = apr,
	year = {2021},
	pages = {801--856},
}

@inproceedings{ellisDreamCoderBootstrappingInductive2021,
	address = {Virtual Canada},
	title = {{DreamCoder}: bootstrapping inductive program synthesis with wake-sleep library learning},
	isbn = {978-1-4503-8391-2},
	shorttitle = {{DreamCoder}},
	url = {https://dl.acm.org/doi/10.1145/3453483.3454080},
	doi = {10.1145/3453483.3454080},
	language = {en},
	urldate = {2024-10-01},
	booktitle = {Proceedings of the 42nd {ACM} {SIGPLAN} {International} {Conference} on {Programming} {Language} {Design} and {Implementation}},
	publisher = {ACM},
	author = {Ellis, Kevin and Wong, Catherine and Nye, Maxwell and Sablé-Meyer, Mathias and Morales, Lucas and Hewitt, Luke and Cary, Luc and Solar-Lezama, Armando and Tenenbaum, Joshua B.},
	month = jun,
	year = {2021},
	pages = {835--850},
}
\bibliographystyle{iclr2026_conference}

\newpage

\appendix

\section{Theoretical Guarantees}
This section analyzes the completeness of the search procedure and compares the computational complexity of this lifted inference approach with that of propositional methods.

\subsection{Lattice-Based Refinement}
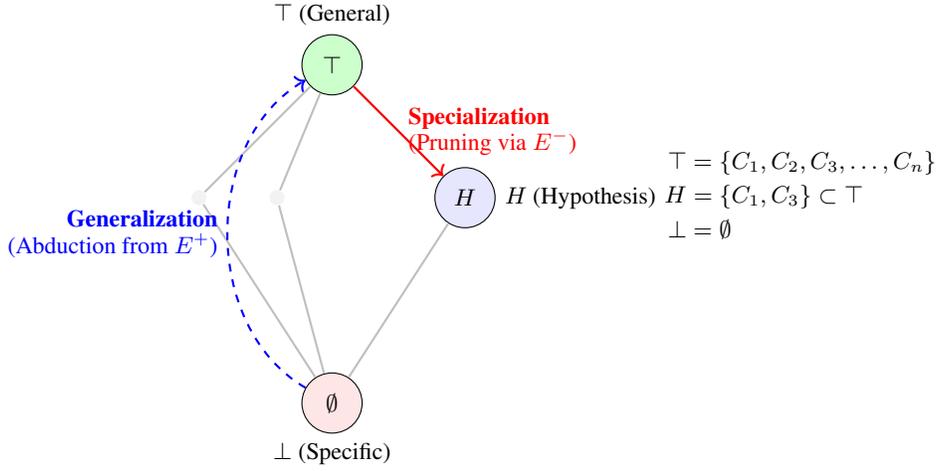
\begin{figure}[h]
\centering
\begin{tikzpicture}[
    node distance=2.5cm,
    every node/.style={font=\small},
    clause/.style={draw, circle, minimum size=0.8cm, inner sep=0pt},
    label_text/.style={align=left, anchor=west}
]

    % --- Nodes ---
    % Bottom (Empty)
    \node[clause, fill=red!10, label=below:{$\bot$ (Specific)}] (bot) {$\emptyset$};

    % Top (Top Program)
    \node[clause, fill=green!20, label=above:{$\top$ (General)}, above of=bot, yshift=2cm] (top) {$\top$};
    
    % Hypothesis (Subset)
    \node[clause, fill=blue!10, label=right:{$H$ (Hypothesis)}, below right of=top] (hyp) {$H$};

    % Ghost nodes to imply the lattice structure
    \node[circle, fill=gray!10, inner sep=2pt, left of=hyp] (ghost1) {};
    \node[circle, fill=gray!10, inner sep=2pt, below left of=top] (ghost2) {};

    % --- Edges (The Lattice) ---
    \draw[thick, gray!50] (bot) -- (ghost1) -- (top);
    \draw[thick, gray!50] (bot) -- (hyp) -- (top);
    \draw[thick, gray!50] (ghost2) -- (top);
    \draw[thick, gray!50] (bot) -- (ghost2);

    % --- The Dynamics (Arrows) ---
    
    % 1. Generalization Arrow (Curved Left)
    \draw[->, thick, blue, dashed] (bot) to[bend left=60] 
        node[midway, left, align=right] {\textbf{Generalization}\\(Abduction from $E^+$)} (top);

    % 2. Specialization Arrow (Straight Down)
    \draw[->, thick, red] (top) -- 
        node[midway, right, align=left] {\textbf{Specialization}\\(Pruning via $E^-$)} (hyp);

    % --- Equations / Legend (Right Side) ---
    \node[label_text, right of=hyp, xshift=2cm] (legend) {
        $\begin{aligned}
            \top &= \{ C_1, C_2, C_3, \dots, C_n \} \\
            H    &= \{ C_1, C_3 \} \subset \top \\
            \bot &= \emptyset
        \end{aligned}$
    };

\end{tikzpicture}
\caption{Lattice traversal dynamics. Prediction errors trigger \textbf{Generalization} (moving to $\top$) or \textbf{Specialization} (pruning to $H$).}
\label{fig:lattice}
\end{figure}

The classification metrics map directly to operators on the subset lattice, see \ref{fig:lattice}:
\begin{itemize}
    \item \textbf{Specialization (Handling $FP$):} When the hypothesis is overly general (predicting events that do not occur), the system prunes the incorrect clauses from the Top Program. This reduces the entailment set, effectively moving down the lattice.
    
    \item \textbf{generalization (Handling $FN$):} When the hypothesis is overly specific (failing to predict observed events), the system invokes the metarule engine to induce (add) new clauses. This expands the entailment set to cover the unexplained observations, effectively moving up the lattice.
\end{itemize}

The search for a hypothesis $H$ corresponds to a traversal of the refinement graph defined by the metarules $\mathcal{M}$ and background knowledge $\mathcal{B}$. We define the expressible hypothesis space $\mathcal{H}_{\mathcal{M}, d}$ as the set of all logic programs derivable by composing metarules from $\mathcal{M}$ up to a maximum derivation depth $d$.

\subsection{Completeness and Expressivity}

\noindent \textbf{Proposition (Completeness):} \textit{If there exists a target hypothesis $H^* \in \mathcal{H}_{\mathcal{M}, d}$ that correctly entails the observed state transitions, the algorithm is guaranteed to find it.}

\noindent \textbf{Proof:} The proof follows the properties of Top Program Construction established in \citep{patsantzisTopProgramConstruction2021}, adapted to our depth-bounded search:
\begin{enumerate}
\item \textbf{Completeness of generalization (Moving Up):} The \texttt{AbduceChain} procedure exhaustively generates all metarule instantiations that entail the current observation $O$ within depth $d$. If the target hypothesis $H^*$ exists in $\mathcal{H}_{\mathcal{M}, d}$, then every clause $C \in H^*$ is technically derivable. Consequently, during the generalization phase, all clauses constituting $H^*$ are added to the initial Top Program $\top_{init}$. Thus, $H^* \subseteq \top_{init}$.
\item \textbf{Soundness of specialization (Moving Down):} The refinement phase iterates through $\top_{init}$ and removes any clause $C$ such that $C \wedge \mathcal{B} \models E^-$. By definition, the target hypothesis $H^*$ is consistent with the environment, meaning no clause in $H^*$ entails a negative example. Therefore, the pruning operator never removes a clause belonging to $H^*$.

\item \textbf{Convergence:} Since $H^*$ is captured during generalization and preserved during Specialization, the final hypothesis $\top_{final}$ satisfies $H^* \subseteq \top_{final}$. Because $H^*$ entails the positive examples, $\top_{final}$ necessarily entails them as well.
\end{enumerate}\hfill $\square$

\section{Algorithms}
\begin{algorithm}
\caption{Self-Supervised Online MIL Loop}
\label{alg:online_mil}
\SetKwInOut{Input}{Input}
\SetKwInOut{Output}{Output}

\Input{Stream of states $S_0, S_1, \dots$, Metarules $\mathcal{M}$, Background Knowledge $\mathcal{B}$}
\Output{Evolving Hypothesis $H = \langle Abs, Dyn, Con \rangle$}

$H \leftarrow \langle \emptyset, \emptyset, \emptyset \rangle$\;
$Context \leftarrow S_0$\;

\For{$t \leftarrow 1$ \KwTo $\infty$}{
    Receive $S_t$\;
    \tcp{Derive Examples via Set Difference}
    $E^+ \leftarrow S_t \setminus S_{t-1}$ \tcp*{Observed Additions}
    $E^- \leftarrow S_{t-1} \setminus S_t$ \tcp*{Observed Removals}
    
    \tcp{Phase 1: Predict (using compiled H)}
    $P_{add} \leftarrow \text{Predict}(H, Context)$\;
    $P_{rem} \leftarrow \text{DeriveRemovals}(P_{add}, H)$\;
    
    \tcp{Phase 2: Verify and Calculate Error}
    $FP_{add} \leftarrow P_{add} \setminus E^+$\;
    $FN_{add} \leftarrow E^+ \setminus P_{add}$\;
    $FP_{rem} \leftarrow P_{rem} \setminus E^-$\;
    $FN_{rem} \leftarrow E^- \setminus P_{rem}$\;
    
    \tcp{Phase 3: Refine}
    \If{$FP_{add} \neq \emptyset$}{
        $Dyn \leftarrow \text{Prune}(Dyn, FP_{add})$\;
    }

    \If{$FP_{rem} \neq \emptyset$}{
        $Con \leftarrow \text{Prune}(Con, FP_{rem})$\;
    }
    
    \If{$FN_{add} \neq \emptyset$}{
        $\langle \Delta Dyn, \Delta Abs \rangle \leftarrow \text{MetaruleInduction}(FN_{add}, Context, \mathcal{M}, \mathcal{B})$\;
        $Dyn \leftarrow Dyn \cup \Delta Dyn$; $Abs \leftarrow Abs \cup \Delta Abs$\;
    }
    \If{$FN_{rem} \neq \emptyset$}{
        $\langle \Delta Con, \Delta Abs' \rangle \leftarrow \text{MetaruleInduction}(FN_{rem}, Context, \mathcal{M}, \mathcal{B})$\;
        $Con \leftarrow Con \cup \Delta Con$; $Abs \leftarrow Abs \cup \Delta Abs'$\;
    }
    
    \tcp{Maintenance}
    $H \leftarrow \text{CompressAndGC}(H)$\;
    $Context \leftarrow S_t$\;
}
\end{algorithm}

\end{document}